# Extraction of Symbolic Rules from Artificial Neural Networks

S. M. Kamruzzaman, and Md. Monirul Islam

*Abstract*—Although backpropagation ANNs generally predict better than decision trees do for pattern classification problems, they are often regarded as black boxes, i.e., their predictions cannot be explained as those of decision trees. In many applications, it is desirable to extract knowledge from trained ANNs for the users to gain a better understanding of how the networks solve the problems. A new rule extraction algorithm, called rule extraction from artificial neural networks (REANN) is proposed and implemented to extract symbolic rules from ANNs. A standard three-layer feedforward ANN is the basis of the algorithm. A four-phase training algorithm is proposed for backpropagation learning. Explicitness of the extracted rules is supported by comparing them to the symbolic rules generated by other methods. Extracted rules are comparable with other methods in terms of number of rules, average number of conditions for a rule, and predictive accuracy. Extensive experimental studies on several benchmarks classification problems, such as breast cancer, iris, diabetes, and season classification problems, demonstrate the effectiveness of the proposed approach with good generalization ability.

## I. INTRODUCTION

THE last two decades have seen a growing number of researchers and practitioners applying ANNs for classification in a variety of real world applications. In some of these applications, it may be desirable to have a set of rules that explains the classification process of a trained network [11]. The classification concept represented as rules is certainly more comprehensible to a human user than a collection of ANNs weights [10].

While the predictive accuracy obtained by ANNs is often higher than that of other methods or human experts, it is generally difficult to understand how the network arrives at a particular conclusion due to the complexity of the ANNs architectures [7]. It is often said that an ANN is practically a "black box". Even for a network with only a single hidden layer, it is generally impossible to explain why a certain pattern is classified as a member of one class and another pattern as a member of another class [9].

Lack of explanation capability is one of the most important reasons why ANNs do not get the necessary interest in the industry. It is therefore necessary that an ANN should be able to explain itself. This can be done in several ways: extracting if-then rules, converting ANNs to decision trees are some of them.

S. M. Kamruzzaman is with the Department of Computer Science and Engineering, Manarat International University, Bangladesh (e-mail: smzaman@gmail.com, smk.cse@manarat.ac.bd).

Md. Monirul Islam is with the Department of Computer Science and Engineering, Bangladesh University of Engineering and Technology (BUET), Bangladesh.

Extracting if-then rules is usually accepted as the best way of extracting the knowledge represented in the ANN. Not because it is an easy job, but because the rules created at the end are more understandable for humans than any other representation [6].

This paper proposes a new rule extraction algorithm, called rule extraction from artificial neural networks (REANN) to extract symbolic rules from ANNs. A standard three-layer feedforward ANN is the basis of the algorithm. A four-phase training algorithm is proposed for backpropagation learning. In the first phase, the number of hidden nodes of the network is determined automatically in a constructive fashion by adding nodes one after another based on the performance of the network on training data. In the second phase, the ANN is pruned such that irrelevant connections and input nodes are removed while its predictive accuracy is still maintained. In the third phase, the continuous activation values of the hidden nodes are discretized by using an efficient heuristic clustering algorithm. And finally in the fourth phase, rules are extracted by examining the discretized activation values of the hidden nodes using a rule extraction algorithm, REx.

## II. RELATED WORKS

There is quite a lot of literature on algorithms that extracts rules from trained ANNs [1] [2]. Several approaches have been developed for extracting rules from a trained ANN. Saito and Nakano [3] proposed a medical diagnosis expert system based on a multiplayer ANN. They treated the network as black box and used it only to observe the effects on the network output caused by change the inputs.

H. Liu and S. T. Tan [4] proposes X2R, a simple and fast algorithm that can applied to both numeric and discrete data, and generate rules from datasets. It can generate perfect rules in the sense that the error rate of the rules is not worse than the inconsistency rate found in the original data. The rules generated by X2R, are order sensitive, i.e, the rules should be fired in sequence.

R. Setiono and H. Liu [5] presents a novel way to understand an ANN. Understanding an ANN is achieved by extracting rules with a three phase algorithm: first, a weight decay backpropagation network is built so that important connections are reflected by their bigger weights; second, the network is pruned such that insignificant connections are deleted while its predictive accuracy is still maintained; and last, rules are extracted by recursively discretizing the hidden node activation values.

R. Setiono [7] proposes a rule extraction algorithm for extracting rules from pruned ANNs for breast cancer diagnosis. The author describes how the activation values of a hidden node can be clustered such that only a finite and usually small number of discrete values need to be considered while at the same time maintaining the network accuracy.

R. Setiono proposes a rule extraction algorithm named NeuroRule [8]. This algorithm extracts symbolic classification rule from a pruned network with a single hidden layer in two steps. First, rules that explain the network outputs are generated in terms of the discretized activation values of the hidden units. Second, rules that explain the discretized hidden unit activation values are generated in terms of the network inputs. When two sets of rules are merged, a DNF representation of network classification is obtained.

Ismail Taha and Joydeep Ghosh [9] propose three rule extraction techniques for knowledge Based Neural Network (KBNN) hybrid systems and present their implementation results. The suitability of each technique depends on the network type, input nature, complexity, the application nature, and the requirement transparency level. The first proposed approach (BIO-RE) is categorized as Black-box Rule Extraction (BRE) technique, while the second (Partial-RE) and third techniques (Full-RE) belong to Link Rule Extraction (LRE) category.

R. Setiono [10] proposes a rule extraction (RX) algorithm to extract rules from a pruned ANN. The process of extracting rules from a trained ANN can be made much easier if the complexity of the ANN has first been removed.

R. Setiono [11] presents MofN3, a new method for extracting M-of-N rules from ANNs. Given a hidden node of a trained ANN with N incoming connections, show how the value of M can be easily computed. In order to facilitate the process of extracting M-of-N rules, the attributes of the dataset have binary values −1 or 1.

R. Setiono, W. K. Leow and Jack M. Zurada [12] describes a method called rule extraction from function approximating neural networks (REFANN) for extracting rules from trained ANNs for nonlinear regression. It is shown that REFAANN produces rules that are almost as accurate as the original networks from which the rules are extracted.

### III. OBJECTIVE OF THE RESEARCH

This paper proposes a hybrid approach with both constructive and pruning components for automatic determination of simplified ANN architectures. The objective of the research are summarized as follows:

i) To develop an efficient algorithm for extracting symbolic rules from ANNs for medical diagnosis problem to explain the functionality of ANNs.
ii) To find an efficient method for clustering the outputs of hidden nodes.
iii) To extract concise rules with high predictive accuracy.

### IV. PROPOSED ALGORITHM

Extracting symbolic rules from trained ANN is one of the promising areas that are commonly used to explain the functionality of ANNs. The aim of this section is to introduce a new algorithm to extract symbolic rules from trained ANNs. The new algorithm is known as rule extraction from ANNs (REANN). Detailed descriptions of REANN are presented below.

*A. The REANN Algorithm*

A standard three-layer feedforward ANN is the basis of the proposed algorithm REANN. The major steps of REANN are summarized in Fig. 1 which are explained further as follows:

**Step 1** Create an initial ANN architecture. The initial architecture has three layers, i.e. an input, an output, and a hidden layer. Initially, the hidden layer contains only one node. The number of nodes in the hidden layer is automatically determined by using a basic constructive algorithm. Randomly initialize all connection weights within a certain small range.

**Step 2** Remove redundant input nodes, and connections between input nodes and hidden nodes and between hidden nodes and output nodes by using a basic pruning algorithm. When pruning is completed, the ANN architecture contains only important nodes and connections. This architecture is saved for the next step.

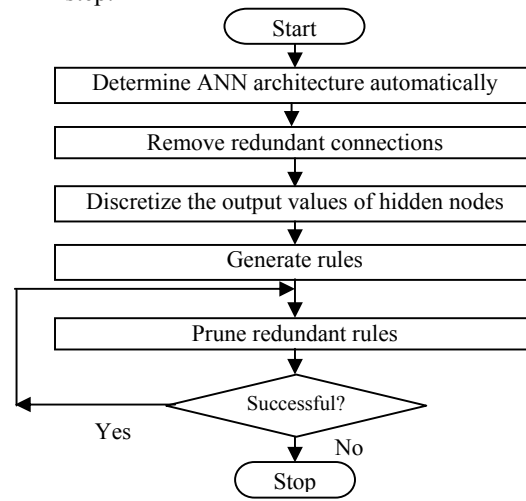

Fig. 1 Flow chart of the REANN algorithm

**Step 3** Discretize the outputs of hidden nodes by using an efficient heuristic clustering algorithm. The reason for discretization is that the outputs of hidden nodes are continuous, thus rules are not readily extractable from the ANN.

**Step 4** Generate rules that map the inputs and outputs relationships.

**Step 5** Prune redundant rules generated in Step 4. Replace specific rules with more general ones.

**Step 6** Check the classification accuracy of the network. If the accuracy falls below an acceptable level, i.e. rule pruning is not successful then stop. Otherwise go to Step 5.

The rules extracted by REANN are compact and comprehensible, and do not involve any weight values. The accuracy of the rules from pruned networks is high as the accuracy of the original networks. The important features of REANN are the rule generated by REx is recursive in nature and is order insensitive, i.e, the rules need not be required to fire sequentially.

*B. Heuristic Clustering Algorithm*

The process of grouping a set of physical or abstract objects into classes of similar objects is called clustering. A cluster of a data objects can be treated collectively as one group in many applications [14]. There exist a large number of clustering algorithms in the literature such as k-means, k-medoids [15] [16]. It is found that some hidden nodes of an ANN maintain almost constant output while other nodes change continuously during the whole training process [17]. Fig. 2 shows a hidden node maintains almost constant output after some training epochs. In REANN, no clustering algorithm is used when hidden nodes maintain almost constant output. If the outputs of hidden nodes do not maintain constant value, a heuristic clustering algorithm is used.

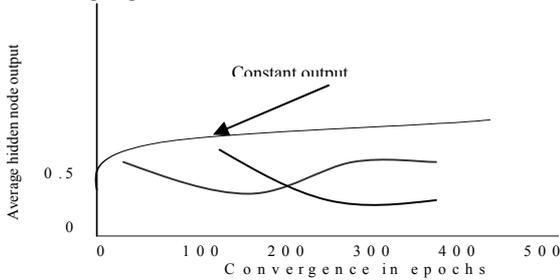

Fig. 2 Output of hidden nodes

The aim of the clustering algorithm is to discretize the output values of hidden nodes. The algorithm places candidates for discrete values such that the distance between them is at least a threshold value ε. The steps of the heuristic clustering algorithm are summarized in Fig. 3, which are explained further as follows:

**Step 1** Let $\varepsilon \in (0, 1)$. D is the activation values in the hidden node. $\delta_1$ is the activation value for the first pattern. The first cluster, $H(1) = \delta_1$, count = 1, and sum(1) = $\delta_1$, set D = 1.

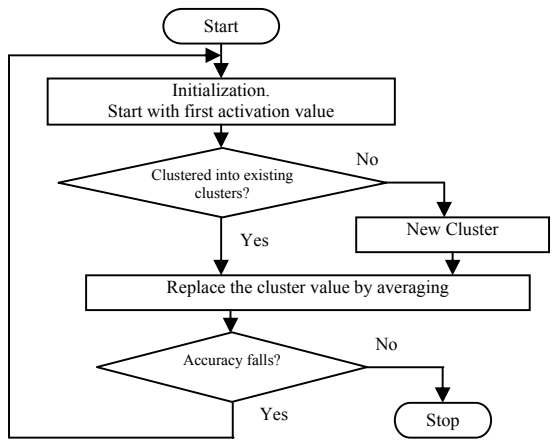

Fig. 3 Flow chart of the heuristic clustering algorithm

**Step 2** For each pattern $p_i$ i = 1, 2, 3, …..k. Checks whether subsequent activation values can be clustered into one of the existing clusters. The distance between an activation value under consideration and its nearest cluster, $|\delta - H(\overline{j})|$, is computed. If this distance is less than ε, then the activation value is clustered in cluster $\overline{j}$. Otherwise, this activation value forms a new cluster. Let δ be its activation value. If there exists an index $\overline{j}$ such that

$$|\delta - H(\overline{j})| = \min_{j \in \{1,2,…D\}} |\delta - H(\overline{j})| \text{ and } |\delta - H(\overline{j})| \leq \varepsilon$$

then set count($\overline{j}$) := count($\overline{j}$)+1,
sum($\overline{j}$) := sum($\overline{j}$)+ δ else D = D+1
H(D) = δ, count(D) = 1, sum (D) = δ.

**Step 3** Replace H by the average of all activation values that have been clustered into this cluster: H(j) := sum(j)/count(j), j=1, 2, 3,…..D.

**Step 4** Once the activation values of all hidden nodes have been obtained, the accuracy of the network is checked with the activation values at the hidden nodes replaced by their discretized values. An activation value δ is replaced by $H(\overline{j})$, where index $\overline{j}$ is chosen such that $\overline{j} = \arg\min_j |\delta - H(j)|$. If the accuracy of the network falls below the required accuracy, then ε must be decreased and the clustering algorithm is run again, otherwise stop.

For a sufficiently small ε, it is always possible to maintain the accuracy of the network with continuous activation values, although the resulting number of different discrete activations can be impractically large.

### D. Rule Extraction Algorithm (REx)

Classification rules are sought in many areas from automatic knowledge acquisition [18] [19] to data mining [20] [21] and ANN rule extraction [22]. The steps of the Rule Extraction (REx) algorithm are summarized in Fig. 4, which are explained further as follows:

**Step 1** Extract Rule:
i=0; while (data is NOT empty/marked){
generate Ri to cover the current pattern and differentiate it from patterns in other categories;
remove/mark all patterns covered by Ri ; i++}

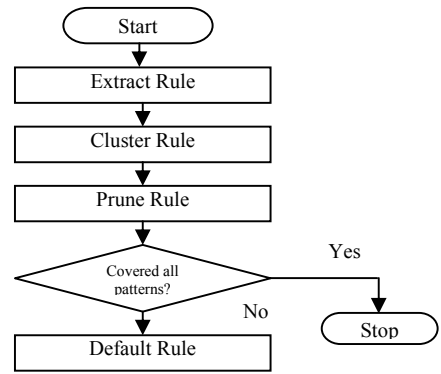

Fig. 4 Flow chart of the rule extraction (REx) algorithm

**Step 2** Cluster Rule:
Cluster rules according to their class levels. Rules generated in Step 1 are grouped in terms of their class levels. In each rule cluster, redundant rules are eliminated; specific rules are replaced by more general rules.

**Step 3** Prune Rule:
replace specific rules with more general ones;
remove noise rules;
eliminate redundant rules;

**Step 4**  Check whether all patterns are covered by any rules. If yes then stop, otherwise continue.
**Step 5**  Determine a default rule:
A default rule is chosen when no rule can be applied to a pattern.

REx exploits the first order information in the data and finds shortest sufficient conditions for a rule of a class that can differentiate it from patterns of other classes. It can generate concise and perfect rules in the sense that the error rate of the rules is not worse than the inconsistency rate found in the original data. The novelty of REx is that the rule generated by it is order insensitive, i.e, the rules need not be required to fire sequentially.

## V. EXPERIMENTAL STUDIES

This section evaluates the performance of REANN on three well-known benchmark classification problems. These are the breast cancer, and iris classification problems.

### A. Data Set Description

The characteristics of the data sets are summarized in Table I. The detailed descriptions of the data sets are available at ics.uci.edu in directory /pub/machine-learning-databases [23] [24].

TABLE I
CHARACTERISTICS OF DATA SETS

| Data Sets | No. of Examples | Input Attributes | Output Classes |
|---|---|---|---|
| Breast Cancer | 699 | 9 | 2 |
| Iris | 150 | 4 | 3 |
| Diabetes | 768 | 8 | 2 |
| Season | 11 | 3 | 4 |

### B. Experimental Setup

In all experiments, one bias node with a fixed input 1 was used for hidden and output layers. The learning rate was set between [0.1, 1.0] and the weights were initialized to random values between [-1.0, 1.0]. Hyperbolic tangent function is used as hidden node activation function and logistic sigmoid function as output node activation function.

In this study, all data sets representing the problems are divided into two sets. One is the training set and the other is the testing set. The numbers of examples in the training set and testing set are based on numbers in other works, in order to make comparison with those works possible. The sizes of the training and testing data sets used in this study are given as follows:
*Breast cancer data set:* the first 350 examples are used for the training set and the rest 349 for the testing set.
*Iris data set*: the first 75 examples are used for the training set and the rest 75 for the testing set.
*Diabetes data set:* the first 384 examples are used for the training set and the rest 384 for the testing set.

### C. Experimental Results

Tables II-V show ANN architectures produced by REANN and training epochs over 10 independent runs on three benchmark classification problems. The initial architecture was selected before applying the constructive algorithm, which was used to determine the number of nodes in the hidden layer. The intermediate architecture was the outcome of the constructive algorithm, and the final architecture was the outcome of pruning algorithm used in REANN.

It is seen that REANN can automatically determine compact ANN architectures. For example, for the breast cancer data, REANN produces more compact architecture. The average number of nodes and connections were 6.8 and 5.8 respectively; in most of the 10 runs 5 to 6 input nodes were pruned.

Fig. 5 shows the smallest of the pruned networks over 10 runs for breast cancer problem. The accuracy of this network on the training data and testing data were 96.275% and 93.429% respectively. In this example only three input attributes $A_1$, $A_6$ and $A_9$ were important and only three discrete values of hidden node activation's were needed to maintain the accuracy of the network.

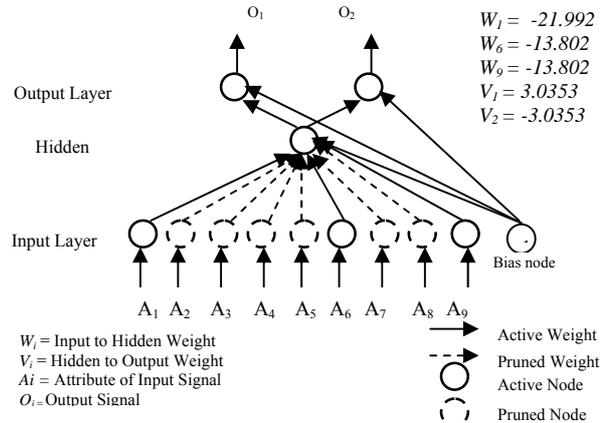

Fig. 5 A pruned network for breast cancer problem

The discrete values found by the heuristic clustering algorithm were 0.987, -0.986 and 0.004. Of the 350 training data, 238 patters have the first value, 106 have the second value and rest 6 patterns have third value. The weight of the connection from the hidden node to the first output node was 3.0354 and to the second output node was –3.0354.

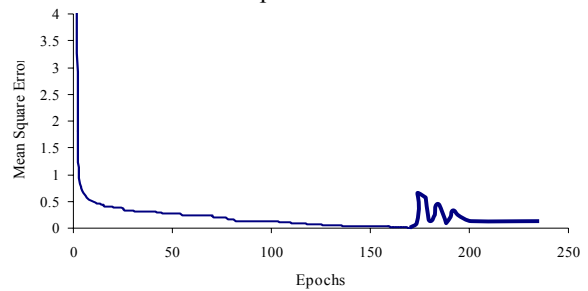

Fig. 6 Training time error for breast cancer problem

Figs. 6 shows the training time error for breast cancer problem. It was observed that the training error decreased and maintained almost constant for a long time after some training epochs and then fluctuates. The fluctuation was made due to the pruning process. As the network was retrained after completing the pruning process thus the training error again maintained almost constant value.

*C.1 Extracted Rules*

The number of rules extracted by REANN and the accuracy of the rules in training and testing data sets were described in Table VI. But the visualization of the rules in terms of the original attributes ware not discussed. The following subsections discussed the rules extracted by REANN in terms of the original attributes. The number of conditions per rule and the number of rules extracted were also visualized here.

*C.1.1 Breast Cancer Data*

Rule 1: If Clump thickness ($A_1$) <= 0.6 and Bare nuclei ($A_6$) <= 0.5 and Mitosis ($A_9$) <= 0.3, then benign
Default Rule: malignant.

*C.1.2 Iris Data*

Rule 1: If Petal-length (A3) <= 1.9 then Iris setosa
Rule 2: If Petal-length (A3) <= 4.9 and Petal-width (A4) <= 1.6 then Iris versicolor
Default Rule: Iris virginica.

*C.1.3 Diabetes Data*

Rule 1: If Plasma glucose concentration ($A_2$) <= 0.64 and Age ($A_8$) <= 0.69
then tested negative
Default Rule: tested positive.

*C.1.4 Season Data*

Rule 1: If Tree (A2) = yellow then autumn
Rule 2: If Tree (A2) = leafless then autumn
Rule 3: If Temperature (A3) = low then winter
Rule 4: If Temperature (A3) = high then summer
Default Rule: spring.

TABLE VI
NUMBER OF EXTRACTED RULES AND RULES ACCURACIES

| Data Sets | No. of Extracted Rules | Rules Accuracy on Training Set | Rules Accuracy on Testing Set |
|---|---|---|---|
| Breast Cancer | 2 | 93.43 % | 96.28 % |
| Iris | 3 | 98.67 % | 97.33 % |
| Diabetes | 2 | 72.14 % | 76.56 % |
| Season | 4 | 100 % | 100 % |

TABLE II
ANN ARCHITECTURES AND TRAINING EPOCHS FOR **BREAST CANCER** DATA. THE RESULTS WERE AVERAGED OVER 10 INDEPENDENT RUNS

|  | Initial Architecture | | Intermediate Architecture | | Final Architecture | | No. of Epoch |
|---|---|---|---|---|---|---|---|
|  | No. of Node | No. of Connection | No. of Node | No. of Connection | No. of Node | No. of Connection |  |
| Mean | 12 (9-1-2) | 11 | 12.7 | 18.1 | 6.8 | 5.8 | 233.2 |
| Min | 12 (9-1-2) | 11 | 12 | 11 | 5 | 5 | 222 |
| Max | 12 (9-1-2) | 11 | 14 | 33 | 10 | 9 | 245 |

TABLE III ANN ARCHITECTURES AND TRAINING EPOCHS FOR **IRIS** DATA. THE RESULTS WERE AVERAGED OVER 10 INDEPENDENT RUNS

|  | Initial Architecture | | Intermediate Architecture | | Final Architecture | | No. of Epoch |
|---|---|---|---|---|---|---|---|
|  | No. of Node | No. of Connection | No. of Node | No. of Connection | No. of Node | No. of Connection |  |
| Mean | 8 (4-1-3) | 7 | 9 | 14 | 8.8 | 10.2 | 196.7 |
| Min | 8 (4-1-3) | 7 | 8 | 7 | 8 | 7 | 183 |
| Max | 8 (4-1-3) | 7 | 10 | 21 | 10 | 14 | 217 |

TABLE IV ANN ARCHITECTURES AND TRAINING EPOCHS FOR **DIABETES** DATA. THE RESULTS WERE AVERAGED OVER 10 INDEPENDENT RUNS

|  | Initial Architecture | | Intermediate Architecture | | Final Architecture | | No. of Epoch |
|---|---|---|---|---|---|---|---|
|  | No. of Node | No. of Connection | No. of Node | No. of Connection | No. of Node | No. of Connection |  |
| Mean | 11 (8-1-2) | 10 | 13.2 | 30 | 12.5 | 19.4 | 302.6 |
| Min | 11 (8-1-2) | 10 | 12 | 20 | 12 | 14 | 279 |
| Max | 11 (8-1-2) | 10 | 14 | 40 | 13 | 24 | 326 |

TABLE V ANN ARCHITECTURES AND TRAINING EPOCHS FOR **SEASON** DATA. THE RESULTS WERE AVERAGED OVER 10 INDEPENDENT RUNS

|  | Initial Architecture | | Intermediate Architecture | | Final Architecture | | No. of Epoch |
|---|---|---|---|---|---|---|---|
|  | No. of Node | No. of Connection | No. of Node | No. of Connection | No. of Node | No. of Connection |  |
| Mean | 8 (3-1-4) | 7 | 8.9 | 13.3 | 8.7 | 11.2 | 88.2 |
| Min | 8 (3-1-4) | 7 | 8 | 7 | 8 | 9 | 73 |
| Max | 8 (3-1-4) | 7 | 10 | 14 | 10 | 16 | 101 |

TABLE VII PERFORMANCE COMPARISON OF REANN WITH OTHER ALGORITHMS FOR **BREAST CANCER** DATA

| Data Set | Feature | REANN | NN RULES | DT RULES | C4.5 | NN-C4.5 | OC1 | CART |
|---|---|---|---|---|---|---|---|---|
| Breast Cancer | No. of Rules | 2 | 4 | 7 | - | - | - | - |
|  | Avg. No. of Conditions | 3 | 3 | 1.75 | - | - | - | - |
|  | Accuracy % | 96.28 | 96 | 95.5 | 95.3 | 96.1 | 94.99 | 94.71 |

TABLE VIII PERFORMANCE COMPARISON OF REANN WITH OTHER ALGORITHMS FOR **IRIS** DATA

| Data Set | Feature | REANN | NN RULES | DT RULES | BIO RE | Partial RE | Full RE |
|---|---|---|---|---|---|---|---|
| Iris | No. of Rules | 3 | 3 | 4 | 4 | 6 | 3 |
|  | Avg. No. of Conditions | 1 | 1 | 1 | 3 | 3 | 2 |
|  | Accuracy % | 98.67 | 97.33 | 94.67 | 78.67 | 78.67 | 97.33 |

TABLE IX PERFORMANCE COMPARISON OF REANN WITH OTHER ALGORITHMS FOR **DIABETES** DATA

| Data Set | Feature | REANN | NN RULES | C4.5 | NN-C4.5 | OC1 | CART |
|---|---|---|---|---|---|---|---|
| Diabetes | No. of Rules | 2 | 4 | - | - | - | - |
| | Avg. No. of Conditions | 2 | 3 | - | - | - | - |
| | Accuracy % | 76.56 | 76.32 | 70.9 | 76.4 | 72.4 | 72.4 |

TABLE X PERFORMANCE COMPARISON OF REANN WITH OTHER ALGORITHMS FOR **SEASONS** DATA

| Data set | Feature | REANN | RULES | X2R |
|---|---|---|---|---|
| Season | No. of Rules | 5 | 7 | 6 |
| | Avg. No. of Conditions | 1 | 2 | 1 |
| | Accuracy % | 100.0 | 100.0 | 100.0 |

Table VI shows number of extracted rules and rules accuracy for six benchmark problems. In most of the cases REANN produces fewer rules with better accuracy.

It was observed that two to three rules were sufficient to solve the problems. The accuracy was 100% for lenses data set for having lower number of examples.

## VI. COMPARISON

This section compares experimental results of REANN with the results of other works. The primary aim of this work is not to exhaustively compare REANN with all other works, but to evaluate REANN in order to gain a deeper understanding of rule extraction.

Table VII compares REANN results of breast cancer problem with those produced by NN RULES [8], DT RULES [8], C4.5 [19], NN-C4.5 [13], OC1 [13], and CART [25] algorithms. REANN achieved best performance although NN RULES was closest second. But number of rules extracted by REANN are 2 whereas these were 4 for NN RULES.

Table VIII compares REANN results of iris data with those produced by NN RULES, DT RULES, BIO RE [9], Partial RE [9], and Full RE [9] algorithms. REANN achieved 98.67% accuracy although NN RULES was closest second with 97.33% accuracy. Here number of rules extracted by REANN and NN RULES are equal.

Table IX compares REANN results of diabetes data with those produced by NN RULES, C4.5, NN-C4.5, OC1, and CART algorithms. REANN achieved 76.56% accuracy although NN-C4.5 was closest second with 76.4% accuracy. Due to the high noise level, the diabetes problem is one of the most challenging problems in our experiments. REANN has outperformed all other algorithms.

Table X compares REANN results of season data with those produced by RULES [26] and X2R [4]. All three algorithms achieved 100% accuracy. This is possible because the number of examples is low. Number of extracted rules by REANN are 5 whereas these were 7 for RULES and 6 for X2R.

## VII. CONCLUSIONS

ANNs are often viewed as black boxes. While their predictive accuracy is high, one usually cannot understand why a particular outcome is predicted due to the complexity of the network. This work is an attempted to open up these black boxes by extracting symbolic rules from it through the proposed efficient rule extraction algorithm REANN.

An important feature of rule extraction algorithm, REx, is its recursive nature. They are concise, comprehensible, order insensitive and do not involve any weight values. The accuracy of the rules from a pruned network is as high as the accuracy of the fully connected network.

Extensive experiments have been carried out in this study to evaluate how well REANN performed on four benchmark classification problems in ANNs including breast cancer, iris, diabetes, and season in comparison with other algorithms. In almost all cases, REANN outperformed the others. With the rules extracted by the method introduced here, ANNs should no longer be regarded as black boxes.

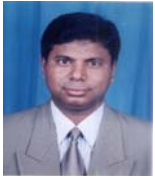

**S. M. Kamruzzaman** received the B. Sc. Engineering degree in Electrical and Electronic Engineering from Bangladesh Institute of Technology (BIT), Dkaka, Bangladesh, in 1997, the M. Sc. Engineering degree in Computer Science and Engineering from Bangladesh University of Engineering and Technology (BUET), Dhaka, Bangladesh, in 2005. From 1998 to 2004, he was a Lecturer and Assistant Professor with the Department of Computer Science and Engineering, International Islamic University Chittagong (IIUC), Chittagong, Bangladesh. In 2005, he moved to Manarat International University, Dhaka, Bangladesh as an Assistant Professor in the Department of Computer Science and Engineering. His research interests include neural networks, data mining, bangla language processing and pattern recognition.